\def\year{2021}\relax
\documentclass[letterpaper]{article} 
\usepackage{aaai21}  
\usepackage{times}  
\usepackage{helvet} 
\usepackage{courier}  
\usepackage[hyphens]{url}  
\usepackage{graphicx} 
\usepackage{natbib}  
\usepackage{caption} 
\usepackage{amsfonts}
\usepackage{amsmath}
\usepackage{multirow}
\usepackage{booktabs}
\usepackage{subfigure}
\usepackage[switch]{lineno}
\title{PointINet: Point Cloud Frame Interpolation Network}
\author{

    Fan Lu\textsuperscript{\rm 1},
    Guang Chen\textsuperscript{\rm 1}\thanks{Guang Chen is the corresponding author.},
    Sanqing Qu\textsuperscript{\rm 1},
    Zhijun Li\textsuperscript{\rm 2},
    Yinlong Liu\textsuperscript{\rm 3},
    Alois Knoll\textsuperscript{\rm 3}
    \\
}
\affiliations{

    \textsuperscript{\rm 1} Tongji University, \textsuperscript{\rm 2} University of Science and Technology of China,
    \textsuperscript{\rm 3} Technische Universität München \\

    \{lufan, guangchen, 2011444\}@tongji.edu.cn,
    zjli@ieee.org, Yinlong.Liu@tum.de, knoll@in.tum.de \\

}

\begin{document}
\maketitle

\begin{abstract}
LiDAR point cloud streams are usually sparse in time dimension, which is limited by hardware performance. Generally, the frame rates of  mechanical LiDAR sensors are 10 to 20 Hz, which is much lower than other commonly used sensors like cameras. To overcome the temporal limitations of LiDAR sensors, a novel task named Point Cloud Frame Interpolation is studied in this paper. Given two consecutive point cloud frames, Point Cloud Frame Interpolation aims to generate intermediate frame(s) between them. To achieve that, we propose a novel framework, namely Point Cloud Frame Interpolation Network (PointINet). Based on the proposed method, the low frame rate point cloud streams can be upsampled to higher frame rates. We start by estimating bi-directional 3D scene flow between the two point clouds and then warp them to the given time step based on the 3D scene flow. To fuse the two warped frames and generate intermediate point cloud(s), we propose a novel learning-based points fusion module, which simultaneously takes two warped point clouds into consideration. We design both quantitative and qualitative experiments to evaluate the performance of the point cloud frame interpolation method and extensive experiments on two large scale outdoor LiDAR datasets demonstrate the effectiveness of the proposed PointINet. Our code is available at \url{https://github.com/ispc-lab/PointINet.git}.
\end{abstract}

\section{Introduction}

\begin{figure}[t]
    \centering
    \includegraphics[width=0.45\textwidth]{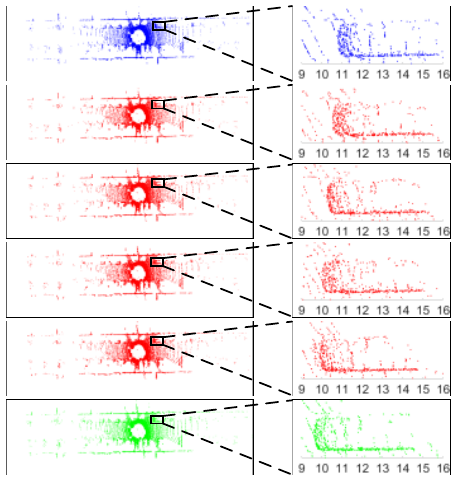}
    \caption{Illustration of Point Cloud Frame Interpolation. The blue and green point clouds are two input frames and the red point clouds are four interpolated frames. We zoom in an area to display the details for better visualization.}
    \label{fig:interpolation}
\end{figure}

LiDAR is one of the most important sensors in numerous applications (\emph{e.g.}, autonomous vehicles and intelligent robots). However, the frame rates of typical mechanical LiDAR sensors (\emph{e.g.}, Velodyne HDL-64E, Hesai Pandar64, etc.) are greatly limited by hardware performance. Frame rates of LiDAR are generally 10 to 20 Hz, which contributes to temporal and spatial discontinuity of point cloud streams. Compared with the low frame rate of LiDAR, the frame rates of other commonly used sensors on intelligent vehicles and robots are typically much higher. For example, the frame rate of cameras and Inertial Measurement Unit (IMU) can achieve over 100 Hz. The large difference in frame rate can cause difficulty to synchronize LiDAR with other sensors. Upsampling low frame rate LiDAR point cloud streams to higher frame rates can be an efficient solution to that \cite{liu2020pseudo}. Besides, higher frame rate may enhance the performance of several applications like object tracking \cite{kiani2017need}. It is worth noting that video frame interpolation is commonly utilized to generate high frame rate videos from low frame rate ones \cite{jiang2018super} (\emph{e.g.}, from 30 Hz to 240 Hz). Compared to the success of video frame interpolation, frame interpolation of 3D point clouds has not been well explored. Therefore, it is needed to explore frame interpolation algorithms for 3D point clouds to overcome the temporal limitations of LiDAR sensors.

Based on the above considerations, a novel task named \textit{Point Cloud Frame Interpolation} is studied in this paper. Given two consecutive point clouds, point cloud frame interpolation aims to predict intermediate point cloud frame according to the given time step to form spatially and temporally coherent point cloud streams (see Fig.~\ref{fig:interpolation}). Consequently, low frame rate LiDAR point cloud streams (10 to 20 Hz) can be upsampled to high frame rate ones (50 to 100 Hz) based on point cloud frame interpolation. 

\begin{figure*}
    \centering
    \includegraphics[width=0.99\textwidth]{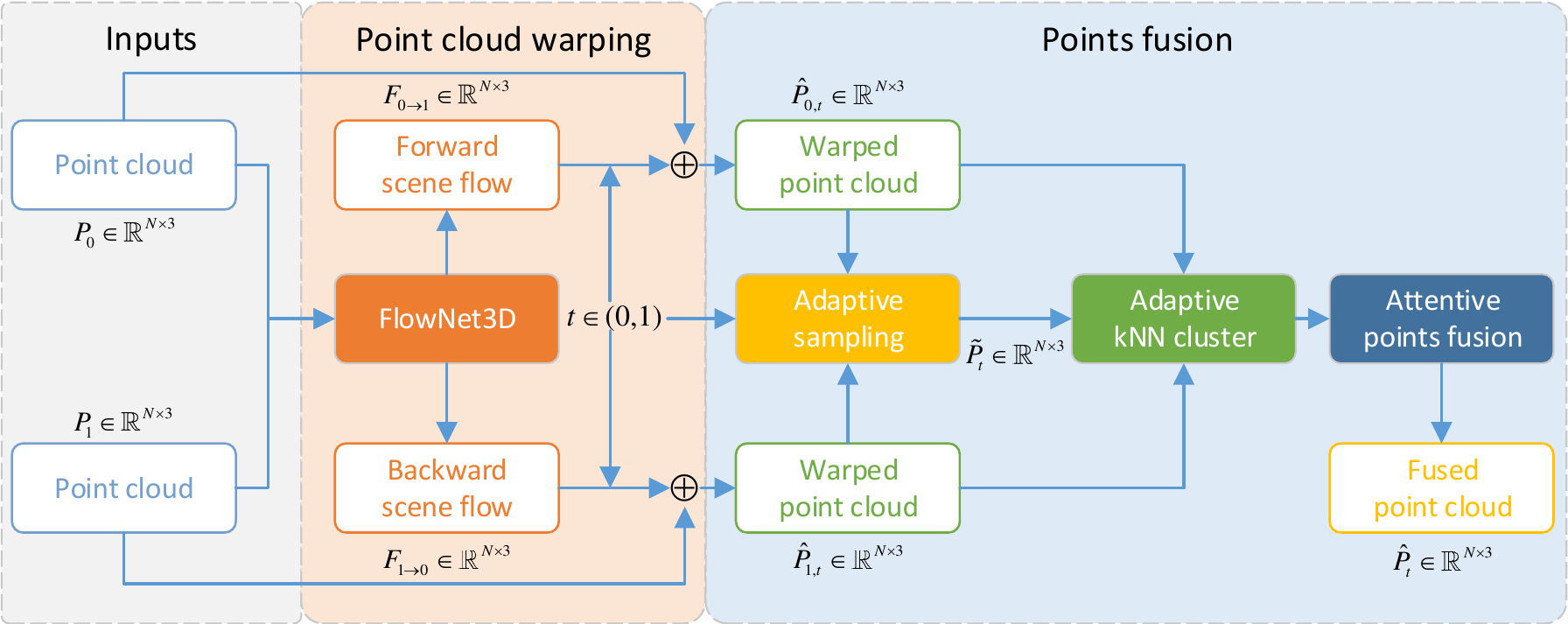}
    \caption{Overall architecture of the proposed PointINet. Given the input two consecutive point clouds, PointINet follows a pipeline consists of point cloud warping module and points fusion module.}
    \label{fig:overall}
\end{figure*}

Concretely, to achieve temporally interpolation of point cloud streams, we proposed a novel learning-based framework named PointINet (Point Cloud Frame Interpolation Network). The proposed PointINet consists of two main components: point cloud warping module and points fusion module. Two consecutive point clouds are firstly input into the point cloud warping module to warp the two point clouds to the given time step. To achieve that, we start by estimating the bi-directional 3D scene flow between two consecutive point clouds for motion estimation. 3D scene flow represents the motion field of points from one point cloud to the other one. Here we adopt a learning-based scene flow estimation network named FlowNet3D \cite{liu2019flownet3d} to predict the 3D scene flow. Then the two point clouds are warped to the given time step based on the linearly interpolated 3D scene flow. Thereafter, the key problem is how to fuse the two frames to form a new intermediate point cloud. 3D point clouds are unstructured and unordered \cite{qi2017pointnet}. Thus, there are no direct correspondences between points in two point clouds like pixels in two images. Consequently, it is non-trivial to perform fusion of the two point clouds. To address the problem, we propose a novel points fusion module. The points fusion module adaptively sample points from two warped point clouds and construct $k$-nearest-neighbor ($k$NN) cluster for each sampled point according to the time step to adjust the contributions of two point clouds. After that, the proposed attentive points fusion adopts an attention mechanism to aggregate points in each cluster to generate the intermediate point clouds. The overall architecture of the proposed PointINet is shown in Fig.~\ref{fig:overall}.

To evaluate the proposed method, we design both qualitative and quantitative experiments. Besides, experiments on applications are also performed to evaluate the quality of the generated interpolated point clouds. Extensive experiments on two large scale outdoor LiDAR datasets demonstrate the effectiveness of the proposed PointINet. 

To summarize, our main contributions are as follows:
\begin{itemize}
    \item To overcome the temporal limitations of LiDAR sensors, a novel task Point Cloud Frame Interpolation is studied.
    \item A new learning-based framework named PointINet is presented to effectively generate intermediate frames between two consecutive point clouds.
    \item Both qualitative and quantitative experiments are conducted to verify the validity of the proposed method.
\end{itemize}

\section{Related work}
In this section we briefly review the literature relevant to point cloud frame interpolation. We start by describing common methods for video frame interpolation and then review 3D scene flow estimation methods for point clouds.
\subsection{Video frame interpolation}
Currently a large number of video frame interpolation methods are based on optical flow estimation \cite{liu2019deep, reda2019unsupervised, jiang2018super, xu2019quadratic, liu2017video}. One of the most representative work of optical flow-based methods is Super SloMo \cite{jiang2018super}, which utilizes learning-based method to predict bi-directional optical flow to estimate the motion between consecutive frames. Then two input frames are further warped and fused with occlusion reasoning to generate the final intermediate frames. \cite{reda2019unsupervised} utilizes cycle consistency to support unsupervised learning of video frame interpolation. \cite{xu2019quadratic} proposes an quadratic video interpolation method to exploit the acceleration information in videos. Another part of the methods for video frame interpolation are kernel-based \cite{niklaus2017videocvpr, niklaus2017videoiccv}. \cite{niklaus2017videocvpr} estimates a kernel on each location and predict the output pixel locations by performing convolution on the patches. \cite{niklaus2017videoiccv} further improves the method by formulating frame interpolation as local separable convolution over input frames using pairs of 1D kernels. Recently, \cite{bao2019memc} combines kernel and optical flow-based methods. They utilize optical flow to predict rough locations of pixels and then refine the location using estimated kernels.

\subsection{3D Scene flow estimation}
3D scene flow of point clouds can be considered as a promotion of 2D optical flow in 3D scenes, which represents the 3D motion field of points. Compared with the high research interest in 2D optical flow estimation \cite{ilg2017flownet, dosovitskiy2015flownet, sun2018pwc}, there is relative little work on 3D scene flow estimation. FlowNet3D \cite{liu2019flownet3d} is a pioneering work of deep learning-based 3D scene flow estimation. \cite{liu2019flownet3d} proposes a flow embedding layer to model the motion of points in different point clouds. Following FlowNet3D, FlowNet3D++ \cite{wang2020flownet3d++} proposes geometric constraints to further improve the accuracy. HPLFlowNet \cite{gu2019hplflownet} introduces Bilateral Convolutional Layers (BCL) in scene flow estimation. PointPWC-Net \cite{wu2019pointpwc} proposes a novel cost volume and estimates the 3D scene flow in a coarse-to-fine manner. Recently, \cite{mittal2020just} provides several unsupervised loss functions to support the generalization of pre-trained scene flow estimation models on more real datasets. In our implementation, we select FlowNet3D to perform 3D scene flow estimation between two point clouds due to the simplicity and effectiveness.

\section{Point cloud frame interpolation}
In this section, we first introduce the overall architecture of the proposed point cloud frame interpolation network (PointINet) and then explain the details of the two key components of PointINet, namely point cloud warping module and points fusion module. 
\subsection{Overall architecture}
The overall architecture of PointINet is shown in Fig.~\ref{fig:overall}. Given two consecutive point clouds $P_0\in \mathbb{R}^{N\times 3}$ and $P_1 \in \mathbb{R}^{N\times 3}$ with a time step $t\in (0,1)$, the goal of PointINet is to predict the intermediate point cloud $\hat{P}_t$ at time step $t$. PointINet consists of two key modules: point cloud warping module to warp the two input point clouds to the given time step $t$ and points fusion module to fuse the two warped point clouds. We will describe the two modules in detail below.

\subsection{Point cloud warping}
Given two point clouds $P_0$ and $P_1$, point cloud warping module aims to predict the position of each point of $P_0$ in $\hat{P}_{0,t}$, where $\hat{P}_{0,t}$ is the corresponding point cloud of $P_0$ at time step $t$ (also predict $\hat{P}_{1,t}$ for $P_1$). The key point here is to estimate the motion of each point from $P_0$ to $\hat{P}_{0,t}$. We first predict the bi-directional 3D scene flow $F_{0\rightarrow 1}\in \mathbb{R}^{N\times 3}$ and $F_{1\rightarrow 0} \in \mathbb{R}^{N\times 3}$ between two point clouds $P_0$ and $P_1$ to estimate the motion of points. 3D scene flow is the 3D motion field of points, which can be regarded as a promotion of optical flow in 3D point clouds. Here we utilize an existing learning-based framework FlowNet3D \cite{liu2019flownet3d} to estimate the bi-directional 3D scene flow. Suppose that the motion of points between two consecutive frames of point clouds is linear, the scene flow $F_{0\rightarrow t}$ and $F_{1\rightarrow t}$ can be approximated by linearly interpolating $F_{0\rightarrow 1}$ and $F_{1\rightarrow 0}$, which can be represented as
\begin{equation}
\begin{split}
    F_{0\rightarrow t} &= t \times F_{0\rightarrow 1} \\
    F_{1\rightarrow t} &= (1-t) \times F_{1\rightarrow 0}
\end{split}
\end{equation}

Then $P_0$ and $P_1$ can be warped to the given time step $t$ based on the interpolated 3D scene flow $F_{0\rightarrow t}$ and ${F_{1\rightarrow t}}$,
\begin{equation}
\begin{split}
    \hat{P}_{0,t} &= P_0 + F_{0\rightarrow t}\\
    \hat{P}_{1,t} &= P_1 + F_{1\rightarrow t}
\end{split}
\end{equation}

\subsection{Points fusion}
The goal of the points fusion module is to fuse the two warped point clouds and generate intermediate point clouds. The architecture of the points fusion module is displayed in the right column of Fig.~\ref{fig:overall}. The input of this module is two warped point clouds $\hat{P}_{0,t}\in\mathbb{R}^{N\times3}$ and $\hat{P}_{1,t}\in\mathbb{R}^{N\times3}$ and the output is the fused intermediate point cloud $\hat{P}_t\in\mathbb{R}^{N\times 3}$. In video frame interpolation, the fusion step mostly concentrates on occlusion and missing regions prediction due to the structured 2D grid-based representation. However, the fusion of two point clouds is non-trivial because point clouds are unstructured and unordered. In the proposed PointINet, we start the fusion by adaptively sampling points from two warped point clouds based on time step $t$ and then construct $k$-nearest-neighbor ($k$NN) clusters centered on the sampled points. After that, the attentive points fusion module adopts an attention mechanism to generate the final intermediate point clouds. The details of the key components of the points fusion module will be described below.

\subsubsection{Adaptive sampling}The first step of the points fusion module is to combine the two warped point clouds to a new point cloud. Intuitively, the contributions of the two point clouds to the intermediate point clouds are not always the same. For example, the intermediate frame $\hat{P}_t$ at $t=0.2$ should be more similar to the first frame $P_0$ than the second frame $P_1$. Based on the above observation, we randomly sample $N_0$ and $N_1$ points from $\hat{P}_{0,t}$ and $\hat{P}_{1,t}$ to generate two sampled point clouds $\tilde{P}_{0,t}\in \mathbb{R}^{N_0\times 3}$ and $\tilde{P}_{1,t} \in \mathbb{R}^{N_1\times 3}$, respectively. Here, $N_0 = (1-t)\times N$ and $N_1 = t\times N$. This operation enables the network to adaptively adjust the contributions of the two warped point clouds according to the target time step $t$. The point cloud close to time step $t$ contributes more to the intermediate frame $\hat{P}_{t}$. After that, $\tilde{P}_{0,t}$ and $\tilde{P}_{1,t}$ are combined to a new point cloud $\tilde{P}_{t}\in \mathbb{R}^{N\times 3}$.

\subsubsection{Adaptive $k$NN cluster}We input $\tilde{P}_{t}$ into the adaptive $k$NN cluster module to generate $k$-nearest-neighbor clusters as input to the followed attentive points fusion module. For each point in $\tilde{P}_{t}$, we search for $K$ nearest neighbors in two warped point clouds $\hat{P}_{0,t}$ and $\hat{P}_{1,t}$. Similar to adaptive sampling, the number of neighbors in $\hat{P}_{0,t}$ and $\hat{P}_{1,t}$ are also adaptively adjusted according to $t$ to balance the contributions of two point clouds. Thus, we query $K_0$ neighbors in $\hat{P}_{0,t}$ and $K_1$ neighbors in $\hat{P}_{1,t}$, where $K_0 = (1-t)\times K$ and $K_1 = t\times K$. As a result, we obtain $N$ clusters and each cluster consists of $K$ neighbor points. Denoting the center point of a cluster as $x^i$ and the neighbor points as $\{x^i_1,\cdots,x^i_k,\cdots,x^i_K\}\in \mathbb{R}^{K\times 3}$. Then each neighbor point is subtracted by the center point as $(x^i_k-x^i)$ to obtain the relative position of neighbor points in a cluster. Besides, the Euclidean distance between neighbor point and the center point $\left\|x^i_k-x^i\right\|_2$ is calculated as an additional channel of the cluster. Consequently, the final feature of a single cluster can be denoted as $F^i=\{f^i_1,\cdots,f^i_k,\cdots,f^i_K\}\in\mathbb{R}^{K\times 4}$.

\subsubsection{Attentive points fusion}
\begin{figure}
    \centering
    \includegraphics[width=0.45\textwidth]{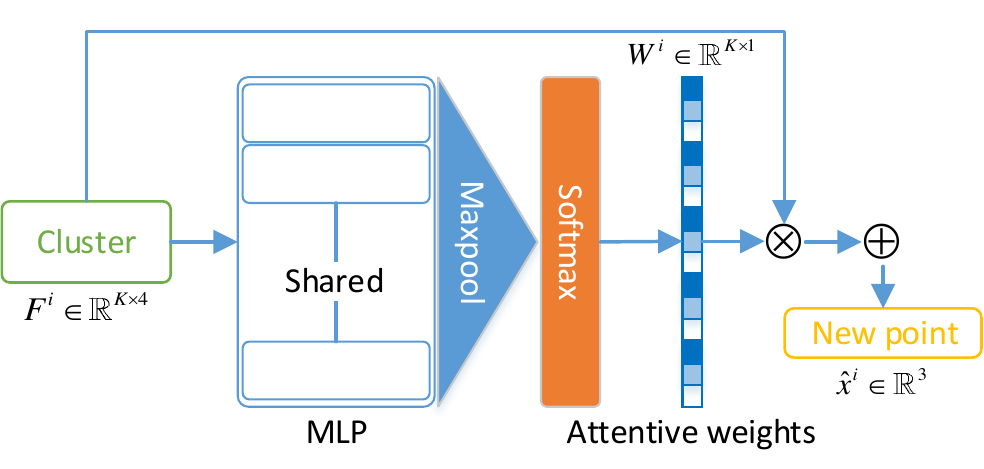}
    \caption{The network architecture of the proposed attentive points fusion module.}
    \label{fig:fusion}
\end{figure}

Attention mechanism has been widely used in 3D point cloud learning \cite{yang2019modeling, wang2019graph, wang2019exploiting}. Here we adopt an attention mechanism to aggregate the feature of neighbor points to generate new points for the intermediate point clouds. The network architecture of the attentive points fusion module can be seen in Fig.~\ref{fig:fusion}. Inspired by PointNet \cite{qi2017pointnet} and PointNet++ \cite{qi2017pointnet++}, we input the feature $F^i$ of a single cluster into a shared multi layer perceptron (Shared-MLP) to generate a feature map. Then the followed maxpool layer and a Softmax function are applied to predict one-dimensional attentive weights $W^i=\{w^i_1,\cdots,w^i_k,\cdots, w^i_K\}\in \mathbb{R}^{K\times 1}$ for all neighbor points in the cluster. After that, the new point $\hat{x}^i$ can be represented as the weighted sum of the neighbor points,
\begin{equation}
    \hat{x}^i = \sum_{k=1}^{K} x^i_k \cdot w^i_k, \quad i=1,\cdots, N
\end{equation}
Finally, the generated intermediate point cloud $\hat{P}_t$ can be represented as $\hat{P}_t=\{\hat{x}^1,\cdots,\hat{x}^N\}\in \mathbb{R}^{N\times 3}$. Intuitively, the proposed attentive points fusion module can assign higher weights to points in the cluster that are more consistent with the target point cloud. After the points fusion module, each generated point in the new intermediate point cloud is aggregated from neighbor points in the two point clouds in its receptive field. Besides, the contributions of two point clouds are dynamically adjusted according to the time step $t$ with the help of adaptive sampling and adaptive $k$NN cluster module. Consequently, the generated intermediate point cloud is an effective fusion of the two input point clouds.

\subsection{Loss}
Chamfer distance \cite{fan2017point} is commonly used to measure the similarity of two point clouds. Here we utilize chamfer distance to supervise the training of the proposed PointINet. Given the generated intermediate point cloud $\hat{P}_t\in\mathbb{R}^{N\times 3}$ and the ground truth one $P_t\in \mathbb{R}^{N\times 3}$, the chamfer distance loss can be represented as
\begin{equation}
    \label{eq:CD}
    \mathcal{L} = \frac{1}{N}\sum_{\hat{x}^i\in \hat{P}_t} \min_{x^j \in P_t} \left\|\hat{x}^i-x^j\right\|_2 + \frac{1}{N}\sum_{x^j\in P_t} \min_{\hat{x}^i \in \hat{P}_t} \left\|\hat{x}^i-x^j\right\|_2
\end{equation}
where $\left\|\cdot\right\|_2$ represents the $L_2$-norm.

\section{Experiments}
We perform both qualitative and quantitative experiments to demonstrate the performance of the proposed method. Besides, we also perform experiments on two applications (\emph{i.e.}, keypoints detection and multi frame Iterative Closest Point (ICP)) to better evaluate the quality of the generated intermediate point clouds.
\subsection{Datasets}
We evaluate the proposed method on two large scale outdoor LiDAR datasets, namely KITTI odometry dataset \cite{geiger2012we} and nuScenes dataset \cite{caesar2020nuscenes}. KITTI odometry dataset provides 11 sequences with ground truth (00-10) and we use sequence 00 to train the network , 01 to validate and the others to evaluate. NuScenes dataset consists of 850 training scenes and we use the first 100 scenes for training and the remaining 750 scenes for evaluation. Due to the lack of high frame rate LiDAR sensors, we simply downsample the 10 Hz point clouds in KITTI odometry dataset to 2 Hz and the 20 Hz point clouds in nuScenes dataset to 4 Hz for training and the quantitative experiments. Consequently, there are 4 intermediate point clouds between two consecutive frames in the downsampled point cloud streams.

\subsection{Implementation details}
We start by training FlowNet3D on Flythings3D dataset \cite{mayer2016large} and then refine the network on KITTI scene flow dataset \cite{menze2015object}. We directly use the data pre-processed by \cite{liu2019flownet3d} to train FlowNet3D. Then we further refine the pre-trained FlowNet3D on KITTI odometry dataset and nuScenes dataset, respectively. During this procedure, the current frame with a randomly selected frame within $N_s$ frames before and after it are used as a training pair. Then the first frame is warped to the second frame with the predicted scene flow and the chamfer distance (see Eq.~\ref{eq:CD}) between the warped point cloud and the second point cloud is adopted as the loss function to supervise the refinement of FlowNet3D. After that, the weight of FlowNet3D is fixed when training the followed points fusion module. During the training of the points fusion module, two consecutive frames and a randomly sampled frame from the 4 intermediate point clouds with the corresponding time step are utilized as a training sample. We randomly downsample the point clouds to 16384 points  during training and the number of neighbor points $K$ is set to 32 in our implementation. The channels of the layers of Shared-MLP in attentive points fusion module are set to $[64,64,128]$. All of the network is implemented using PyTorch \cite{paszke2019pytorch} and Adam is used as the optimizer. Besides, the points fusion module is only trained on KITTI odometry dataset and we simply generalize the trained model to nuScenes dataset for evaluation.

\subsection{Qualitative experiments}
\begin{figure*}[t]
    \centering
    \includegraphics[width=0.99\textwidth]{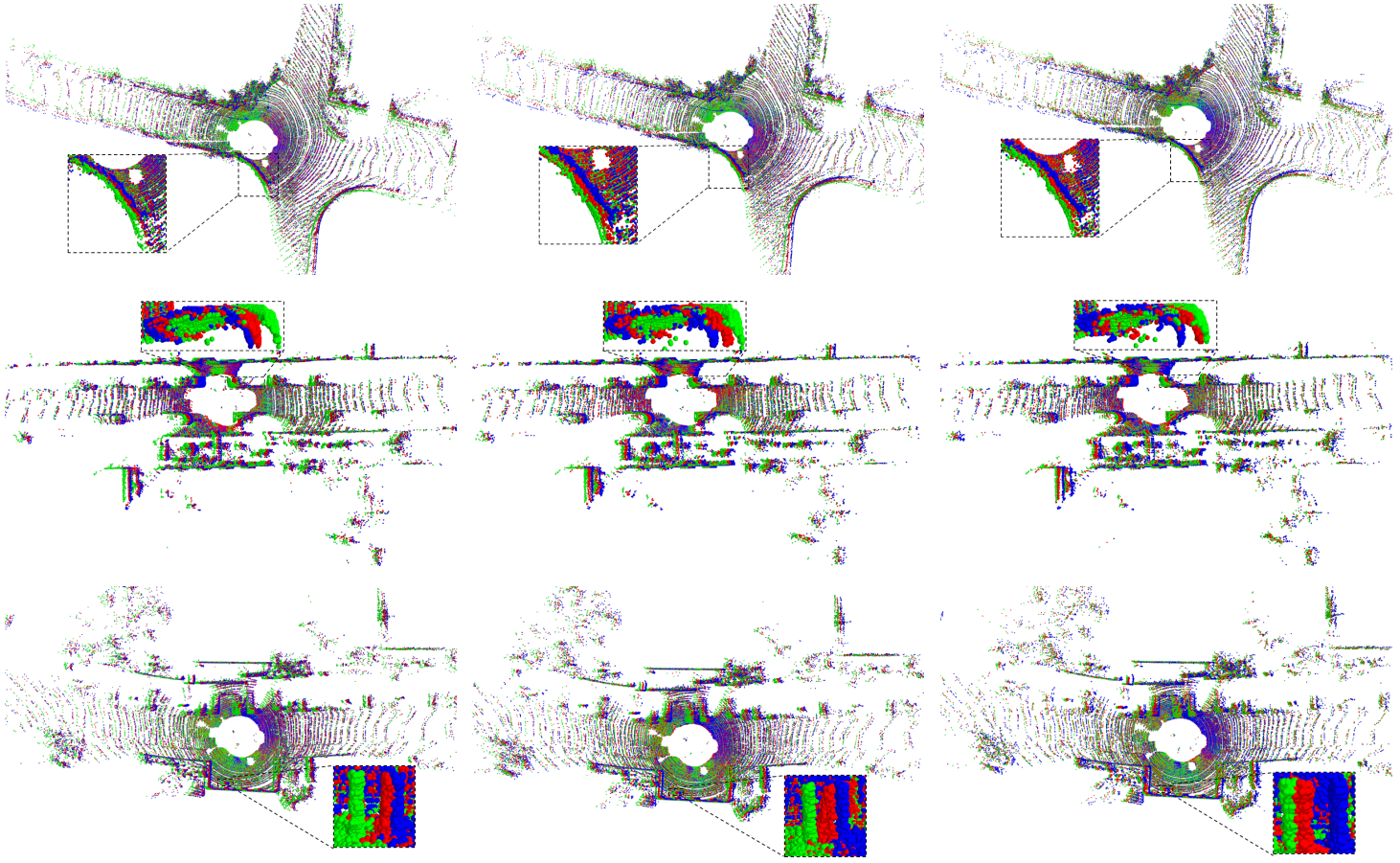}
    \caption{Qualitative results of the proposed PointINet. From top to bottom rows are the interpolation results of 3 pairs of consecutive frames. The time step $t$ of columns from left to right are 0.25, 0.50 and 0.75, respectively. Blue, green and red point clouds represent first frames, second frames and the predicted intermediate frames, respectively. Besides, we zoom in an area of the point cloud and then rotate it to a proper perspective to better visualize the details of the interpolated point cloud.}
    \label{fig:qualitative}
\end{figure*}

The goal of the proposed PointINet is to generate high frame rate LiDAR streams from low frame rate ones. However, there are no existing high frame rate LiDAR sensors. Thus, we train the FlowNet3D with $N_s=1$ to provide proper scene flow estimation for closer point clouds and then directly apply the points fusion module trained on the downsampled point cloud streams on 10 Hz point cloud streams of KITTI odometry dataset to generate high frame rate point cloud streams. Here we provide a qualitative visualization in Fig.~\ref{fig:qualitative}, where the number of points here is set to 32768. The 10 Hz point cloud streams are upsampled to 40 Hz and the time step of intermediate frames are set to 0.25, 0.50 and 0.75. According to Fig.~\ref{fig:qualitative}, the proposed PointINet well estimates the motion of points between two clouds and the fusion algorithm can preserve the details of the point cloud. In addition to that, we also provide several demo videos in the supplementary materials to compare high frame rate point cloud streams with low frame rate point cloud streams. According to the demo videos, the high frame rate point cloud streams are obviously temporally and spatially smoother than low frame rate ones. 

\subsection{Quantitative experiments}
\subsubsection{Evaluation metrics} We evaluate the similarity and consistency between the generated point clouds and the ground truth ones on the downsampled point cloud streams using two evaluation metrics: Chamfer distance (CD) and Earth mover's distance (EMD). CD is previously described in Eq.~\ref{eq:CD}. EMD is also a commonly used metric to compare two point clouds \cite{Weng2020_SPF2}, which is implemented by solving a linear assignment problem. Given two point clouds $\hat{P}_t\in\mathbb{R}^{N\times 3}$ and $P_t\in\mathbb{R}^{N\times 3}$, EMD can be calculated as
 \begin{equation}
     EMD = \min_{\phi : \hat{P}_t\rightarrow P_t} \frac{1}{N}\sum_{\hat{x}\in \hat{P}_t} \left\|\hat{x}-\phi(\hat{x})\right\|_2
 \end{equation}
where $\phi: \hat{P}_t\rightarrow P_t$ is a bijection. 

\subsubsection{Baselines} To demonstrate the performance of the proposed PointINet, we define 3 baselines to make comparison with our method: (1) \textit{Identity}. We simply duplicate the first point cloud frame as the intermediate point clouds. (2) \textit{Align-ICP}. We firstly estimate the rigid transformation between the two consecutive frame of point clouds using Iterative Closest Point (ICP) algorithm and then linearly interpolate that to obtain the transformation between the first frame and intermediate frame. Thereafter, the first point cloud is transformed to the intermediate frame based on the transformation. (3) \textit{Scene flow}. We estimate the 3D scene flow between the consecutive two frames using FlowNet3D and calculate the scene flow from the first frame to the intermediate frame by linear interpolation. Then the intermediate point clouds are obtained by transform the first point cloud according to the 3D scene flow. All of the point clouds are downsampled to 16384 points by randomly sampling in quantitative experiments. 

\subsubsection{Results}The CD and EMD of the proposed PointINet and other baselines on KITTI odometry dataset and nuScenes dataset are shown in Table~\ref{tab:kitti_cd_emd} and Table~\ref{tab:nuscenes_cd_emd}, respectively. According to the results, the performance of our method significantly outperforms other baselines. For example, the chamfer distance of the proposed PointINet is about $1/3$, $3/5$ and $2/3$ of \textit{Identity}, \textit{Align-ICP} and \textit{Scene flow} on KITTI odometry dataset, respectively. It is worth noting that our method is superior to \textit{Scene flow} by an obvious margin, which also reflects the effectiveness of the points fusion module. Noting that we only train the points fusion module on KITTI odometry dataset and the results on nuScenes dataset also demonstrate the generalization ability of the network.
\begin{table}[t]
    \centering
    \begin{tabular}{cccccc}
    \toprule
        Metric & Identity & Align-ICP & Scene flow & Ours \\
        \midrule
        CD$\downarrow$ & 1.398 & 0.752 & 0.687 & \textbf{0.457} \\
        EMD$\downarrow$ & 68.93 & 83.79 & 57.13 & \textbf{39.46}\\
        \bottomrule
    \end{tabular}
    \caption{Results of quantitative evaluation of PointINet and other baselines on KITTI odometry dataset.}
    \label{tab:kitti_cd_emd}
\end{table}

\begin{table}[t]
    \centering
    \begin{tabular}{cccccc}
    \toprule
        Metric & Identity & Align-ICP & Scene flow & Ours \\
        \midrule
        CD$\downarrow$ & 0.617 & 0.555 & 0.511 & \textbf{0.487} \\
        EMD$\downarrow$ & 54.24 & 51.12 & 50.97 & \textbf{47.98}\\
        \bottomrule
    \end{tabular}
    \caption{Results of quantitative evaluation of PointINet and other baselines on nuScenes dataset.}
    \label{tab:nuscenes_cd_emd}
\end{table}
 
\subsection{Applications}
In order to better evaluate the quality of the generated intermediate point clouds and the similarity with original point clouds, we apply two applications on the interpolated point cloud streams and the original ones, namely Keypoints detection and Multi frame ICP. We firstly respectively downsample the 10 Hz point clouds in KITTI odometry dataset and 20 Hz point clouds in nuScenes dataset to 5 Hz and 10 Hz and then interpolate them to the original frame rates as the interpolated point cloud streams. The results of the two applications on the two different point cloud streams are compared to verify the validity of the proposed PointINet.
\subsubsection{Keypoints detection} We perform 3D keypoints detection in the two point cloud streams and evaluate the repeatability of the detected keypoints. We choose 3 handcrafted 3D keypoints, namely SIFT-3D \cite{flint2007thrift}, Harris-3D \cite{sipiran2011harris} and ISS \cite{zhong2009intrinsic}. All of the keypoints are extracted using the implementation in PCL \cite{rusu20113d}. A keypoint in a point cloud is considered repeatable if its distance to the nearest keypoint in the other point cloud (after rigid transformation based on the ground truth pose) is within a threshold $\delta_r$ ($\delta_r$ is set to 0.5 m here) and the repeatability is the ratio of repeatable keypoints. We calculate the average repeatability of keypoints in current point cloud with keypoints in 5 frames before and after it and the number of keypoints is set to 256. Due to the lack of per-frame ground truth pose in nuScenes dataset, the keypoints detection experiments are only performed on KITTI odometry dataset and the results are shown in Table~\ref{tab:repeat_kitti}. According to the results, the repeatability of the interpolated point cloud streams is only slightly reduced compared with the original point cloud streams. For example, the repeatability of Harris-3D of interpolated point clouds is only 0.017 lower than that of original point clouds. The results reflects the high consistency of the generated intermediate point clouds with the ground truth point clouds from the side.

\begin{table}[t]
    \centering
    \begin{tabular}{cccc}
    \toprule
        Keypoints & Harris-3D &  SIFT-3D & ISS \\
        \midrule
        Original & 0.155 & 0.174 & 0.163 \\
        Interpolated & 0.138 & 0.151 & 0.133 \\
        \bottomrule
    \end{tabular}
    \caption{The repeatability of 3 different keypoints of original and interpolated point clouds on KITTI odometry dataset.}
    \label{tab:repeat_kitti}
\end{table}

\subsubsection{Multi frame ICP} We perform iterative closest point (ICP) algorithm on $N_m$ consecutive frames to estimate the rigid transformation between the first and last frames. $N_m$ is set to 10 on KITTI odometry dataset. For nuScenes dataset, the ground truth pose is only provided for keyframes (about 2 Hz). Thus, $N_m$ is set to be the same as the number of frames between two keyframes on nuScenes dataset. We utilize the implementation in PCL to perform ICP algorithm. The one-by-one transformation are accumulated to obtain the transformation between the first and last frames. Relative translation error (RTE) and relative rotation error (RRE) are calculated to evaluate the error of the estimated transformation of multi frame ICP. The results on KITTI odometry dataset and nuScenes dataset are displayed in Table~\ref{tab:icp_kitti} and Table~\ref{tab:icp_nuscenes}, respectively. We also calculate the difference between the errors of original and interpolated point cloud streams and display the results in right column of Table~\ref{tab:icp_kitti} and Table~\ref{tab:icp_nuscenes} for better comparison. According to the results, the RTE and RRE of the multi-frame ICP algorithm on the interpolated point cloud streams are very close to that on original point cloud streams. For example, The RTE on nuScenes dataset of the two results differs by only 0.07 m according to Table~\ref{tab:icp_nuscenes}. The close performance indicates the similarity between the generated intermediate point clouds with the ground truth ones. 

According to the experiments on the two applications, the performance of the algorithm on the interpolated point clouds is slightly inferior to the original point cloud streams due to the possible error of the proposed interpolation method. Nonetheless, the close performance on the two applications proves the high similarity and consistency of the generated point clouds with the original ones. 
\begin{table}[t]
    \centering
    \begin{tabular}{cccc}
    \toprule
    Metric & Original & Interpolated & Difference \\
    \midrule
    RTE (m)& 4.31 & 4.57 & 0.26\\
    RRE (deg) & 2.70 & 2.95 & 0.25\\
    \bottomrule
    \end{tabular}
    \caption{The performance of multi frame ICP of original and interpolated point cloud streams on KITTI odometry dataset.}
    \label{tab:icp_kitti}
\end{table}

\begin{table}[t]
    \centering
    \begin{tabular}{cccc}
    \toprule
    Metric & Original & Interpolated & Difference \\
    \midrule
    RTE (m)& 1.65 & 1.72 & 0.07\\
    RRE (deg)& 0.91 & 0.92 & 0.01\\
    \bottomrule
    \end{tabular}
    \caption{The performance of multi frame ICP of original and interpolated point cloud streams on nuScenes dataset.}
    \label{tab:icp_nuscenes}
\end{table}

\subsection{Efficiency}
The efficiency of the proposed PointINet is evaluated on a PC with NVIDIA Geforce RTX 2060 and the average runtime to generate one intermediate frame for point clouds contain 16384, 32768 and 65536 points are displayed in Table~\ref{tab:runtime}. According to the results, most of the runtime is used to warp the point cloud and the proposed points fusion module requires relatively little time for computation. However, the computation time for points fusion module increases with the number of points due to the per-point computation for fusion. Overall, the proposed PointINet can efficiently generate intermediate frames.

\begin{table}[t]
    \centering
    \begin{tabular}{cccc}
    \toprule
        Number of points & 16384 & 32768 & 65536 \\
    \midrule
        Point cloud warping & 167.3 & 291.1 & 529.3\\
        Points fusion & 36.4 & 81.3 & 196.6 \\
        PointINet & 203.7 & 372.4 & 725.9 \\
    \bottomrule
    \end{tabular}
    \caption{The runtime (ms) of PointINet and its components for different number of points.}
    \label{tab:runtime}
\end{table}

\subsection{Ablation study}
We perform several ablation studies to analyze the effect of different components of the proposed PointINet (\emph{e.g.}, adaptive sampling, adaptive $k$NN cluster and attentive points fusion) to the final results. The experimental setting is consistent with the quantitative experiments and we also use chamfer distance (CD) and earth mover's distance (EMD) to evaluate the performance. All of the ablation studies are performed on KITTI odometry dataset.

\subsubsection{Adaptive sampling} We replace the adaptive sampling strategy by simply randomly sampling half of the points in two warped point clouds to form a new point cloud as the input to the adaptive $k$NN cluster module. The results are shown in the second row of Table~\ref{tab:ablation}. Based on the results, the CD and EMD increase by 0.123 and 8.54 without adaptive sampling, which demonstrates that the adaptive sampling strategy significantly improves the performance. 

\subsubsection{Adaptive $k$NN cluster} We query $K/2$ neighbor points from the two warped point clouds fixedly rather than query points based on time step $t$. According to the results displayed in third row of Table~\ref{tab:ablation}, the CD and EMD without adaptive $k$NN cluster increase from 0.457 to 0.534 and 39.46 to 41.66, respectively. The results prove the effectiveness of the adaptive $k$NN cluster module.

\subsubsection{Attentive points fusion} To demonstrate the effect of the attentive points fusion module, we directly use the point cloud $\tilde{P}_t$ from adaptive sampling as the intermediate point cloud and display the results in the bottom row of Table~\ref{tab:ablation}. According to the results, the attentive points fusion module obviously enhances the final performance.

\begin{table}
    \centering
    \begin{tabular}{ccc}
    \toprule
        Methods & CD$\downarrow$ & EMD$\downarrow$ \\
    \midrule
        full PointINet & \textbf{0.457} & \textbf{39.46} \\
        w/o adaptive sampling & 0.580 & 48.00 \\
        w/o adaptive $k$NN cluster & 0.534 & 41.66 \\
        w/o attentive points fusion & 0.555 & 40.67 \\
    \bottomrule
    \end{tabular}
    \caption{The quantitative evaluation results of ablation studies on KITTI odometry dataset.}
    \label{tab:ablation}
\end{table}

\section{Conclusions}

In this paper, a novel task named  \textit{Point Cloud Frame Interpolation} is studied and a learning-based framework PointINet is designed for this task. Given two consecutive point clouds, the task aims to predict temporally and spatially consistent intermediate frames between them. Consequently, low frame rate point cloud streams can be upsampled to high frame rates using the proposed method. To achieve that, we utilize an existing scene flow estimation network for motion estimation and then warp the two point clouds to the given time step. Then a novel learning-based points fusion module is presented to efficiently fuse the two point clouds. We design both qualitative and quantitative experiments for this task. Extensive experiments on KITTI odometry dataset and nuScenes dataset demonstrate the performance and effectiveness of the proposed PointINet.

\section{Ethics statement}
The proposed point cloud frame interpolation method may have positive effects on the development of autonomous driving and intelligent robots, which can reduce the workload of human drivers and workers and also the incidence of traffic accidents. However, this development may also bring unemployment of human drivers and workers. Besides, the proposed method may have potential military applications like military unmanned aerial vehicles, which can threaten the safety of humans. We should explore more applications which can improve the quality of human life rather than harmful ones.

\section{Acknowledgments}
This work is funded by National Natural Science Foundation of China (No. 61906138), the European Union’s Horizon 2020 Framework Programme for Research and Innovation under the Specific Grant Agreement No. 945539 (Human Brain Project SGA3), and the Shanghai AI Innovation Development Program 2018.

\bibliography{bib.bib}

\end{document}